\documentclass[11pt,a4paper]{article}
\usepackage[hyperref]{naaclhlt2019}
\usepackage{times}
\usepackage{latexsym}
\usepackage{graphicx}
\usepackage{caption}
\usepackage{booktabs}
\usepackage{array, makecell}
\usepackage{url}
\PassOptionsToPackage{hyphens}{url}
\usepackage[hyphenbreaks]{breakurl}
\usepackage{enumerate}
\usepackage{multirow}

\usepackage{pgfplotstable} 
\usepackage{pgfplots}
\pgfplotsset{compat=1.14}
\usetikzlibrary[patterns]

\aclfinalcopy 
\def\aclpaperid{1500} 

\usepackage{amsmath,amsfonts,bm}

\def\eqref#1{equation~\ref{#1}}

\def\1{\bm{1}}

\DeclareMathAlphabet{\mathsfit}{\encodingdefault}{\sfdefault}{m}{sl}
\SetMathAlphabet{\mathsfit}{bold}{\encodingdefault}{\sfdefault}{bx}{n}

\newcommand*\samethanks[1][\value{footnote}]{\footnotemark[#1]}

\def\cnndm{CNN-DailyMail}
\newcommand{\ROUGE}{{\sc Rouge}}

\newcommand{\shared}{{\sc{shared}}}

\newcommand{\srcelmo}{{\sc{src-elmo}}}

\newcommand{\srcelmosharedec}{{\sc{src-elmo+shdemb}}}
\newcommand{\srcft}{{\sc{src-ft}}}

\newcommand{\tgtelmo}{{\sc{tgt-elmo}}}
\newcommand{\tgtft}{{\sc{tgt-ft}}}

\title{Pre-trained Language Model Representations for Language Generation}

\author{Sergey Edunov\thanks{\hspace{0.06in}Equal contribution.} , Alexei Baevski\samethanks{}\hspace{0.05in}, Michael Auli \\
  Facebook AI Research \\
  Menlo Park, CA \\
}

\date{}

\begin{document}
\maketitle
\begin{abstract}
Pre-trained language model representations have been successful in a wide range of language understanding tasks.
In this paper, we examine different strategies to integrate pre-trained representations into sequence to sequence models and apply it to neural machine translation and abstractive summarization.
We find that pre-trained representations are most effective when added to the encoder network which slows inference by only 14\%.
Our experiments in machine translation show gains of up to 5.3 BLEU in a simulated resource-poor setup. 
While returns diminish with more labeled data, we still observe improvements when millions of sentence-pairs are available.
Finally, on abstractive summarization we achieve a new state of the art on the full text version of \cnndm{}.
\footnote{Code and pre-trained models are available at \url{https://github.com/pytorch/fairseq/tree/bi_trans_lm/examples/pretraining}}
\end{abstract}

\section{Introduction}

Pre-training of language models has been shown to provide large improvements for a range of language understanding tasks~\citep{peters2018acl,radford2018unsup,phang2018stilts,devlin2018bert}.
The key idea is to train a large generative model on vast corpora and use the resulting representations on tasks for which only limited amounts of labeled data is available.
Pre-training of sequence to sequence models has been previously investigated for text classification \citep{dai2015arxiv} but not for text generation.
In neural machine translation, there has been work on transferring representations from high-resource language pairs to low-resource settings \citep{zoph2016transfer}.

In this paper, we apply pre-trained representations from language models to language generation tasks that can be modeled by sequence to sequence architectures.
Previous work on integrating language models with sequence to sequence models focused on the decoder network and added language model representations right before the output of the decoder \citep{gulcehre2015using}.
We extend their study by investigating several other strategies such as inputting ELMo-style representations \citep{peters2018acl} or fine-tuning the language model (\textsection\ref{sec:strategies}).

Our experiments rely on strong transformer-based language models trained on up to six billion tokens (\textsection\ref{sec:setup}).
We present a detailed study of various strategies in different simulated labeled training data scenarios and observe the largest improvements in low-resource settings but gains of over 1 BLEU are still possible when five million sentence-pairs are available.
The most successful strategy to integrate pre-trained representations is as input to the encoder network (\textsection\ref{sec:results}).

\section{Strategies to add representations}
\label{sec:strategies}

We consider augmenting a standard sequence to sequence model with pre-trained representations following an ELMo-style regime (\textsection\ref{sec:elmo}) as well as by fine-tuning the language model (\textsection\ref{sec:finetune}).

\subsection{ELMo augmentation}
\label{sec:elmo}

The ELMo approach of \citet{peters2018acl} forms contextualized word embeddings based on language model representations without adjusting the actual language model parameters. 
Specifically, the ELMo module contains a set of parameters $\lambda_1 \dots \lambda_L, \gamma$ to form a linear combination of the $L$ layers of the language model: ELMo = $\gamma \sum_{i=0}^L \frac{1}{Z} \exp(\lambda_i) \mathbf{h}^k$ where $\gamma$ is a learned scalar, $Z$ is a constant to normalize the $\exp(\lambda_i)$ to sum to one and $\mathbf{h}^k$ is the output of the $k$-th language model layer; the module also considers the input word embeddings of the language model. 
We also apply layer normalization \citep{ba2016layer} to each $\mathbf{h}^k$ before computing ELMo vectors.

We experiment with an ELMo module to input contextualized embeddings either to the encoder (\srcelmo{}) or the decoder (\tgtelmo{}).
This provides word representations specific to the current input sentence and these representations have been trained on much more data than is available for the text generation task.

\subsection{Fine-tuning approach}
\label{sec:finetune}

Fine-tuning the pre-trained representations adjusts the language model parameters by the learning signal of the end-task \citep{radford2018unsup,devlin2018bert}.
We replace learned input word embeddings in the encoder network with the output of the language model (\srcft{}).
Specifically, we use the language model representation of the layer before the softmax and feed it to the encoder.
We also add dropout to the language model output.
Tuning separate learning rates for the language model and the sequence to sequence model may lead to better performance but we leave this to future work.
However, we do tune the number of encoder blocks $N$ as we found this important to obtain good accuracy for this setting.
We apply the same strategy to the decoder: we input language model representations to the decoder network and fine-tune the language model when training the sequence to sequence model (\tgtft{}).

\section{Experimental setup}
\label{sec:setup}

\subsection{Datasets}
\label{sec:datasets}

\paragraph{Pre-training.}
We train language models on two languages:
One model is estimated on the German newscrawl distributed by WMT'18 comprising 260M sentences or 6B tokens.
Another model is trained on the English newscrawl data comprising 193M sentences or 5B tokens.
We learn a joint Byte-Pair-Encoding (BPE; Sennrich et al., 2016)\nocite{sennrich:bpe:2016} vocabulary of 37K types on the German and English newscrawl and train the language models with this vocabulary.

\paragraph{Machine translation.}
We consider two benchmarks: 
Most experiments are run on the WMT'18 English-German (en-de) news translation task and we validate our findings on the WMT'18 English-Turkish (en-tr) news task.
For WMT'18 English-German, the training corpus consists of all available bitext excluding the ParaCrawl corpus and we remove sentences longer than 250 tokens as well as sentence-pairs with a source/target length ratio exceeding 1.5.
This results in 5.18M sentence pairs. 
We tokenize all data with the Moses tokenizer~\cite{koehn:moses:2007} and apply the BPE vocabulary learned on the monolingual corpora.

For WMT'18 English-Turkish, we use all of the available bitext comprising 208K sentence-pairs without any filtering.
We develop on newstest2017 and test on newstest2018.
For en-tr we only experiment with adding representations to the encoder and therefore apply the language model vocabulary to the source side. 
For the target vocabulary we learn a BPE code with 32K merge operations on the Turkish side of the bitext.
Both datasets are evaluated in terms of case-sensitive de-tokenized BLEU~\cite{papineni:bleu:2002,post:sacre:2018}.\footnote{sacreBLEU signatures: BLEU+case.mixed+lang.en-\{de,tr\}+numrefs.1+smooth.exp+test.wmt18+tok.13a\\+version.1.2.1}

\paragraph{Summarization.}
We consider the \cnndm{} abstractive document summarization task comprising over 280K news articles paired with multi-sentence summaries. 
\cnndm{} is a widely used dataset for abstractive text summarization.
Following \citep{see17}, we report results on the non-anonymized version of \cnndm{} rather than the entity-anonymized version \citep{hermann15,nallapati16} because the language model was trained on full text.
Articles are truncated to 400 tokens \citep{see17} and we use a BPE vocabulary of 32K types \citep{fan17}. 
We evaluate in terms of F1-Rouge, that is Rouge-1, Rouge-2 and Rouge-L \citep{lin04}.\footnote{We use the following parameters for \texttt{ROUGE-1.5.5.pl}: -m -a -n 2}

\pgfplotstableread[row sep=\\,col sep=&]{
size & share & src-elmo & src-elmo+sharedec & tgt-elmo & tgt-ft & src-ft-noenc & src-ft \\
160K & 2.92 & 3.77 & 5.33 & 0.37 & 2.15 & 2.7 & 2.73 \\
320K & 1.12 & 2.4 & 2.65 & -0.25 & -1.22 & 2.25 & 1.52 \\
640K & 0.7 & 1.95 & 2.28 & -0.3 & -0.84 & 1.48 & 0.6 \\
1280K & 0.55 & 1.54 & 2.02 & -0.01 & -1.11 & 1.14 & 0.77 \\
2560K & 0.18 & 1.48 & 1.55 & 0.21 & -0.95 & 0.78 & 0.6 \\
5186K & -0.03 & 0.97 & 0.90 & -0.3 & -1.13 & 0.2 & -0.03 \\
}\endedata

\begin{figure*}[t]
\begin{tikzpicture}
\begin{axis}[
ybar,
bar width=.15cm,
width=1.0*\textwidth,
height=.48\textwidth,
legend style={at={(0.99,0.83)},
anchor=east,legend columns=3},
xticklabels from table={\endedata}{size},
xtick=data,
ymin=-1.8, ymax=6,
ytick={-1,0,1,2,3,4,5,6},
ylabel={BLEU delta wrt baseline},
xlabel={Bitext tokens}
]
\addplot[black,fill=cyan,postaction={pattern=north east lines}] table[x expr=\coordindex,y=share]{\endedata};
\addplot[black,fill=pink,postaction={pattern=grid}] table[x expr=\coordindex,y=src-elmo]{\endedata};
\addplot[black,fill=brown,postaction={pattern=vertical lines}] table[x expr=\coordindex,y=src-ft-noenc]{\endedata};
\addplot[black,fill=green,postaction={pattern=dots}] table[x expr=\coordindex,y=tgt-elmo]{\endedata};
\addplot[black,fill=yellow,postaction={pattern=north west lines}] table[x expr=\coordindex,y=tgt-ft]{\endedata};
\addplot[black,fill=olive,postaction={pattern=crosshatch}] table[x expr=\coordindex,y=src-elmo+sharedec]{\endedata};
\legend{\shared{},\srcelmo{},\srcft{},\tgtelmo{},\tgtft{},\srcelmosharedec{}}
\end{axis}
\end{tikzpicture}
\caption{
BLEU difference to a bitext-only baseline when adding pre-trained language model representations to a neural machine translation model in different simulated bitext settings.
Results are based on averaging newstest2012-2017 of WMT English-German translation.
}
\label{fig:data_news2017_delta_bar}
\end{figure*}
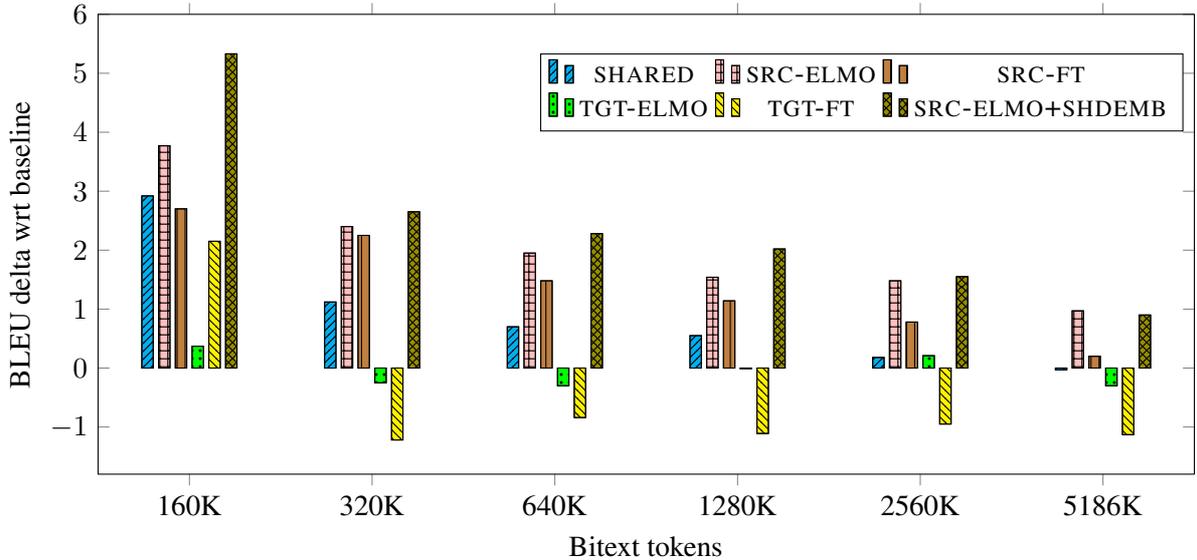

\subsection{Language model pre-training}

We consider two types of architectures: a bi-directional language model to augment the sequence to sequence encoder and a uni-directional model to augment the decoder.
Both use self-attention \citep{vaswani2017transformer} and the uni-directional model contains $N=12$ transformer blocks, followed by a word classifier to predict the next word on the right.
The bi-directional model solves a cloze-style token prediction task at training time~\citep{baevski2019cloze}.
The model consists of two towers, the \emph{forward} tower operates left-to-right and the tower operating right-to-left as \emph{backward} tower; each tower contains $N=12$ transformer blocks.
The forward and backward representations are combined via a self-attention module and the output of this module is used to predict the token at position $i$.
The model has access to the entire input surrounding the current target token.
Models use the standard settings for the Big Transformer \citep{vaswani2017transformer}.
The bi-directional model contains 353M parameters and the uni-directional model 190M parameters.
Both models were trained for 1M steps using Nesterov's accelerated gradient \citep{sutskever2013icml} with momentum $0.99$ following \citet{baevski2018adp}. 
The learning rate is linearly warmed up from $10^{-7}$ to $1$ for 16K steps and then annealed using a cosine learning rate schedule with a single phase to 0.0001 \citep{cosine}.
We train on 32 Nvidia V100 SXM2 GPUs and use the NCCL2 library as well as the torch distributed package for inter-GPU communication.
Training relies on 16-bit floating point operations~\citep{ott:scaling:2018} and it took six days for the bi-directional model and four days for the uni-directional model.

\subsection{Sequence to sequence model}

We use the transformer implementation of the fairseq toolkit~\citep{ott2019fairseq}.
The WMT en-de and en-tr experiments are based on the Big Transformer sequence to sequence architecture with 6 blocks in the encoder and decoder.
For abstractive summarization we use a base transformer model \citep{vaswani2017transformer}.
We tune dropout values of between 0.1 and 0.4 on the validation set.
Models are optimized with Adam \citep{kingma2015adam} using $\beta_1 = 0.9$, $\beta_2 = 0.98$, and $\epsilon = 1e-8$ and we use the same learning rate schedule as \citet{vaswani2017transformer}; we perform 10K-200K depending on bitext size.
All models use label smoothing with a uniform prior distribution over the vocabulary $\epsilon = 0.1$ \citep{szegedy:inception:2015,pereyra:regularize:2017}.
We run experiments on 8 GPUs and generate translations with a beam of size 5.

\section{Results}
\label{sec:results}

\subsection{Machine translation}

We first present a comparison of the various strategies in different simulated parallel corpus size settings.
For each experiment, we tune the dropout applied to the language model representations, and we reduce the number of optimizer steps for smaller bitext setups as models converge faster; all other hyper-parameters are equal between setups.
Our baseline is a Big Transformer model and we also consider a variant where we share token embeddings between the encoder and decoder (\shared; Inan et al., 2016; Press \& Wolf, 2016)\nocite{inan2016tying,press2016using}.

Figure~\ref{fig:data_news2017_delta_bar} shows results averaged over six test sets relative to the baseline which does not share source and target embeddings (Appendix~\ref{app:endeall} shows a detailed breakdown). 
\shared{} performs very well with little labeled data but the gains erode to practically zero in large bitext settings.
Pre-trained language model representations are most effective in low bitext setups.
The best performing strategy is ELMo embeddings input to the encoder (\srcelmo{}). This improves the baseline by 3.8 BLEU in the 160K bitext setting and it still improves the 5.2M setting by over 1 BLEU.

\begin{table}[t]
\centering
\begin{tabular}{lrrr}
\toprule
 & 160K & 640K & 5186K \\ \midrule
baseline & 21.4 & 33.1 & 40.1 \\ \midrule
\srcelmo{} & 26.6 & 35.6 & 41.8 \\
\srcft{} & 24.3 & 34.9 & 40.8 \\
\tgtelmo{} & 21.3 & 31.9 & 40.5 \\
\tgtft{} & 24.2 & 31.4 & 38.8 \\
\srcelmosharedec{} & 29.0 & 36.2 & 41.8 \\
\bottomrule
\end{tabular}
\caption{BLEU on newstest2018 of WMT English-German in three simulated bitext size scenarios.
}
\label{tab:mttest}
\end{table}

\begin{table}[t]
\centering
\begin{tabular}{lrr}
\toprule
& news2017 & news2018 \\ 
\midrule
baseline & 9.8 & 9.5 \\ \midrule
\srcelmo{} & 12.0 & 11.3 \\
\srcelmosharedec{} & 12.9 & 11.8 \\
\bottomrule
\end{tabular}
\caption{WMT English-Turkish translation results in terms of BLEU on newstest2017 (valid) and newstest2018 (test) with ELMo inputs to the encoder.
}
\label{tab:entr}
\end{table}

We further improve \srcelmo{} by sharing learned word representations in the decoder by tying input and output embeddings (\srcelmosharedec{}).
This configuration performs even better than \srcelmo{} with a gain of 5.3 BLEU in the 160K setup.
Sharing decoder embeddings is equally applicable to \srcft{}.
Language model representations are much less effective in the decoder: \tgtft{} improves the 160K bitext setup but yields no improvements thereafter and \tgtelmo{} performs even worse. 
We conjecture that pre-trained representations give much easier wins in the encoder.
Table~\ref{tab:mttest} shows additional results on newstest2018.

Pre-trained representations mostly impacts the training time of the sequence to sequence model (see Appendix~\ref{app:speed}):
\srcelmo{} slows throughput during training by about 5.3x and \srcft{} is even slower because of the need to backpropagate through the LM for fine-tuning (9.2x).
However, inference is only 12-14\% slower than the baseline when adding pre-trained embeddings to the encoder (\srcelmo{}, \srcft{}).
This is because the LM computation can be paralelized for all input tokens.
Inference is much slower when adding representations to the decoder because the LM needs to be invoked repeatedly. 
Our current implementation does not cache LM operations for the previous state and can be made much faster.

The baseline uses a BPE vocabulary estimated on the language model corpora (\textsection\ref{sec:setup}). Appendix~\ref{app:endeall} shows that this vocabulary actually leads to sligtly better performance than a joint BPE code learned on the bitext as is usual.

Next, we validate our findings on the WMT'18 English-Turkish task for which the bitext is truly limited (208K sentence-pairs).
We use the language model vocab for the the English side of the bitext and a BPE vocabulary learned on the Turkish side.
Table~\ref{tab:entr} shows that ELMo embeddings for the encoder improve English-Turkish translation.

\begin{table}[t]
\centering 
\begin{tabular}{lccc}
\toprule
& \multicolumn{3}{c}{\ROUGE{}}\\
& 1 & 2 & L\\    
\toprule
Lead-3 & 40.34 & 17.70 & 36.57 \\
\citet{see17} & 39.53 & 17.28 & 36.38 \\
\citet{gehrmann2018bottom} & 41.22 & 18.68 & 38.34 \\
\midrule
baseline & 40.07 & 17.61 & 36.78 \\
\srcelmosharedec{} & 41.56 & 18.94 & 38.47 \\
\bottomrule
\end{tabular}
\caption{Abstractive summarization results on \cnndm{}. ELMo inputs achieve a new state of the art.}
\label{tab:abs}
\end{table}

\subsection{Abstractive summarization}

Following \citet{see17}, we experiment on the non-anonymized version of \cnndm{}. 
When generating summaries, we follow standard practice of tuning the maximum output length and disallow repeating the same trigram~\citep{paulus17intra,fan17}. 
For this task we train language model representations on the combination of newscrawl and the \cnndm{} training data.
Table~\ref{tab:abs} shows that pre-trained embeddings can significantly improve on top of a strong baseline transformer.
We also compare to \citet{gehrmann2018bottom} who use a task-specific architecture compared to our generic sequence to sequence baseline. 
Pre-trained representations are complementary to their method.

\section{Conclusion}

We presented an analysis of different strategies to add pre-trained language model representations to sequence to sequence models for neural machine translation and abstractive document summarization.
Adding pre-trained representations is very effective for the encoder network and while returns diminish when more labeled data becomes available, we still observe improvements when millions of examples are available.
In future research we will investigate ways to improve the decoder with pre-trained representations.

\bibliography{master}

\begin{thebibliography}{31}
\expandafter\ifx\csname natexlab\endcsname\relax\def\natexlab#1{#1}\fi

\bibitem[{Ba et~al.(2016)Ba, Kiros, and Hinton}]{ba2016layer}
Jimmy~Lei Ba, Jamie~Ryan Kiros, and Geoffrey~E Hinton. 2016.
\newblock Layer normalization.
\newblock \emph{arXiv}, abs/1607.06450.

\bibitem[{Baevski and Auli(2018)}]{baevski2018adp}
Alexei Baevski and Michael Auli. 2018.
\newblock Adaptive input representations for neural language modeling.
\newblock \emph{arXiv}, abs/1809.10853.

\bibitem[{Baevski et~al.(2019)Baevski, Edunov, Liu, Zettlemoyer, and
  Auli}]{baevski2019cloze}
Alexei Baevski, Sergey Edunov, Yinhan Liu, Luke Zettlemoyer, and Michael Auli.
  2019.
\newblock Cloze-driven pretraining of self-attention networks.
\newblock \emph{arXiv}.

\bibitem[{Dai and Le(2015)}]{dai2015arxiv}
Andrew~M. Dai and Quoc~V. Le. 2015.
\newblock Semi-supervised sequence learning.
\newblock \emph{arXiv}, abs/1511.01432.

\bibitem[{Devlin et~al.(2018)Devlin, Chang, Lee, and
  Toutanova}]{devlin2018bert}
Jacob Devlin, Ming-Wei Chang, Kenton Lee, and Kristina Toutanova. 2018.
\newblock Bert: Pre-training of deep bidirectional transformers for language
  understanding.
\newblock \emph{CoRR}, abs/1810.04805.

\bibitem[{Fan et~al.(2017)Fan, Grangier, and Auli}]{fan17}
Angela Fan, David Grangier, and Michael Auli. 2017.
\newblock Controllable abstractive summarization.
\newblock \emph{arXiv}, abs/1711.05217.

\bibitem[{Gehrmann et~al.(2018)Gehrmann, Deng, and Rush}]{gehrmann2018bottom}
Sebastian Gehrmann, Yuntian Deng, and Alexander~M Rush. 2018.
\newblock Bottom-up abstractive summarization.
\newblock \emph{arXiv}, abs/1808.10792.

\bibitem[{Gulcehre et~al.(2015)Gulcehre, Firat, Xu, Cho, Barrault, Lin,
  Bougares, Schwenk, and Bengio}]{gulcehre2015using}
Caglar Gulcehre, Orhan Firat, Kelvin Xu, Kyunghyun Cho, Loic Barrault, Huei-Chi
  Lin, Fethi Bougares, Holger Schwenk, and Yoshua Bengio. 2015.
\newblock On using monolingual corpora in neural machine translation.
\newblock \emph{arXiv}, abs/1503.03535.

\bibitem[{Hermann et~al.(2015)Hermann, Ko\v{c}isk\'{y}, Grefenstette, Espeholt,
  Kay, Suleyman, and Blunsom}]{hermann15}
Karl~Moritz Hermann, Tom\'{a}\v{s} Ko\v{c}isk\'{y}, Edward Grefenstette, Lasse
  Espeholt, Will Kay, Mustafa Suleyman, and Phil Blunsom. 2015.
\newblock Teaching machines to read and comprehend.
\newblock In \emph{Proc. of NIPS}.

\bibitem[{Inan et~al.(2016)Inan, Khosravi, and Socher}]{inan2016tying}
Hakan Inan, Khashayar Khosravi, and Richard Socher. 2016.
\newblock Tying word vectors and word classifiers: {A} loss framework for
  language modeling.
\newblock \emph{arXiv}, abs/1611.01462.

\bibitem[{Kingma and Ba(2015)}]{kingma2015adam}
Diederik~P. Kingma and Jimmy Ba. 2015.
\newblock {Adam: A Method for Stochastic Optimization}.
\newblock In \emph{Proc. of ICLR}.

\bibitem[{Koehn et~al.(2007)Koehn, Hoang, Birch, Callison-Burch, Federico,
  Bertoldi, Cowan, Shen, Moran, Zens, Dyer, Bojar, Constantin, and
  Herbst}]{koehn:moses:2007}
Philipp Koehn, Hieu Hoang, Alexandra Birch, Chris Callison-Burch, Marcello
  Federico, Nicola Bertoldi, Brooke Cowan, Wade Shen, Christine Moran, Richard
  Zens, Chris Dyer, Ondrej Bojar, Alexandra Constantin, and Evan Herbst. 2007.
\newblock Moses: Open source toolkit for statistical machine translation.
\newblock In \emph{Proc. of ACL Demo Session}.

\bibitem[{Lin(2004)}]{lin04}
Chin-Yew Lin. 2004.
\newblock Rouge: A package for automatic evaluation of summaries.
\newblock In \emph{Workshop on Text Summarization Branches Out}.

\bibitem[{Loshchilov and Hutter(2016)}]{cosine}
Ilya Loshchilov and Frank Hutter. 2016.
\newblock {SGDR:} stochastic gradient descent with restarts.
\newblock \emph{arXiv}, abs/1608.03983.

\bibitem[{Nallapati et~al.(2016)Nallapati, Zhou, Gulcehre, Xiang
  et~al.}]{nallapati16}
Ramesh Nallapati, Bowen Zhou, Caglar Gulcehre, Bing Xiang, et~al. 2016.
\newblock Abstractive text summarization using sequence-to-sequence rnns and
  beyond.
\newblock In \emph{Proc. of CONLL}.

\bibitem[{Ott et~al.(2019)Ott, Edunov, Baevski, Fan, Gross, Ng, Grangier, and
  Auli}]{ott2019fairseq}
Myle Ott, Sergey Edunov, Alexei Baevski, Angela Fan, Sam Gross, Nathan Ng,
  David Grangier, and Michael Auli. 2019.
\newblock fairseq: A fast, extensible toolkit for sequence modeling.
\newblock In \emph{Proc. of NAACL System Demonstrations}.

\bibitem[{Ott et~al.(2018)Ott, Edunov, Grangier, and Auli}]{ott:scaling:2018}
Myle Ott, Sergey Edunov, David Grangier, and Michael Auli. 2018.
\newblock Scaling neural machine translation.
\newblock In \emph{Proc. of WMT}.

\bibitem[{Papineni et~al.(2002)Papineni, Roukos, Ward, and
  Zhu}]{papineni:bleu:2002}
Kishore Papineni, Salim Roukos, Todd Ward, and Wei-Jing Zhu. 2002.
\newblock {BLEU}: a method for automatic evaluation of machine translation.
\newblock In \emph{Proc. of ACL}.

\bibitem[{Paulus et~al.(2017)Paulus, Xiong, and Socher}]{paulus17intra}
Romain Paulus, Caiming Xiong, and Richard Socher. 2017.
\newblock A deep reinforced model for abstractive summarization.
\newblock \emph{arXiv}, abs/1705.04304.

\bibitem[{Pereyra et~al.(2017)Pereyra, Tucker, Chorowski, Kaiser, and
  Hinton}]{pereyra:regularize:2017}
Gabriel Pereyra, George Tucker, Jan Chorowski, Lukasz Kaiser, and Geoffrey~E.
  Hinton. 2017.
\newblock Regularizing neural networks by penalizing confident output
  distributions.
\newblock In \emph{International Conference on Learning Representations
  ({ICLR}) Workshop}.

\bibitem[{Peters et~al.(2018)Peters, Neumann, Iyyer, Gardner, Clark, Lee, and
  Zettlemoyer}]{peters2018acl}
Matthew~E Peters, Mark Neumann, Mohit Iyyer, Matt Gardner, Christopher Clark,
  Kenton Lee, and Luke Zettlemoyer. 2018.
\newblock Deep contextualized word representations.
\newblock In \emph{Proc. of ACL}.

\bibitem[{Phang et~al.(2018)Phang, Fevry, and Bowman}]{phang2018stilts}
Jason Phang, Thibault Fevry, and Samuel~R. Bowman. 2018.
\newblock Sentence encoders on stilts: Supplementary training on intermediate
  labeled-data tasks.
\newblock \emph{arXiv}, abs/1811.01088.

\bibitem[{Post(2018)}]{post:sacre:2018}
Matt Post. 2018.
\newblock A call for clarity in reporting bleu scores.
\newblock \emph{arXiv}, abs/1804.08771.

\bibitem[{Press and Wolf(2017)}]{press2016using}
Ofir Press and Lior Wolf. 2017.
\newblock Using the output embedding to improve language models.
\newblock In \emph{Proc. of EACL}.

\bibitem[{Radford et~al.(2018)Radford, Narasimhan, Salimans, and
  Sutskever}]{radford2018unsup}
Alec Radford, Karthik Narasimhan, Tim Salimans, and Ilya Sutskever. 2018.
\newblock Improving language understanding by generative pre-training.
\newblock
  \url{https://s3-us-west-2.amazonaws.com/openai-assets/research-covers/language-unsupervised/language_understanding_paper.pdf}.

\bibitem[{See et~al.(2017)See, Liu, and Manning}]{see17}
Abigail See, Peter~J Liu, and Christopher~D Manning. 2017.
\newblock Get to the point: Summarization with pointer-generator networks.
\newblock In \emph{Proc. of ACL}.

\bibitem[{Sennrich et~al.(2016)Sennrich, Haddow, and Birch}]{sennrich:bpe:2016}
Rico Sennrich, Barry Haddow, and Alexandra Birch. 2016.
\newblock Neural machine translation of rare words with subword units.
\newblock In \emph{Proc. of ACL}.

\bibitem[{Sutskever et~al.(2013)Sutskever, Martens, Dahl, and
  Hinton}]{sutskever2013icml}
Ilya Sutskever, James Martens, George~E. Dahl, and Geoffrey~E. Hinton. 2013.
\newblock {On the importance of initialization and momentum in deep learning}.
\newblock In \emph{Proc. of ICML}.

\bibitem[{Szegedy et~al.(2015)Szegedy, Vanhoucke, Ioffe, Shlens, and
  Wojna}]{szegedy:inception:2015}
Christian Szegedy, Vincent Vanhoucke, Sergey Ioffe, Jonathon Shlens, and
  Zbigniew Wojna. 2015.
\newblock {Rethinking the Inception Architecture for Computer Vision}.
\newblock \emph{arXiv preprint arXiv:1512.00567}.

\bibitem[{Vaswani et~al.(2017)Vaswani, Shazeer, Parmar, Uszkoreit, Jones,
  Gomez, Kaiser, and Polosukhin}]{vaswani2017transformer}
Ashish Vaswani, Noam Shazeer, Niki Parmar, Jakob Uszkoreit, Llion Jones,
  Aidan~N. Gomez, Lukasz Kaiser, and Illia Polosukhin. 2017.
\newblock {Attention Is All You Need}.
\newblock In \emph{Proc. of NIPS}.

\bibitem[{Zoph et~al.(2016)Zoph, Yuret, May, and Knight}]{zoph2016transfer}
Barret Zoph, Deniz Yuret, Jonathan May, and Kevin Knight. 2016.
\newblock Transfer learning for low-resource neural machine translation.
\newblock In \emph{Proc. of EMNLP}.

\end{thebibliography}
\bibliographystyle{acl_natbib}

\clearpage
\onecolumn
\appendix
\section{Detailed WMT English-German Results}
\label{app:endeall}

\begin{table*}[h!]
\begin{small}
\centering
\begin{tabular}{llrrrrrrrrr}
\toprule
bitext & method & \bf 2012 & \bf 2013 & \bf 2014 & \bf 2015 & \bf 2016 & \bf 2017 & \bf 2018 & \bf Avg \\ \midrule
\multirow{8}{*}{\shortstack{160K}}
 & baseline & 13.2 & 15.7 & 13.5 & 15.7 & 18.6 & 14.8 & 21.4 & 16.1 & \\
 & \shared{} & 15.3 & 18.2 & 16.7 & 19.0 & 21.6 & 18.2 & 24.9 & 19.1 & \\
 & \shared{}+bitext-BPE & 15.1 & 17.9 & 16.2 & 18.9 & 22.0 & 18.0 & 25.2 & 19.0 & \\
 & \srcelmo{} & 16.0 & 19.4 & 17.1 & 19.9 & 23.0 & 18.7 & 26.6 & 20.1 & \\
 & \srcft{} & 15.3 & 18.5 & 16.6 & 18.9 & 20.8 & 17.6 & 24.3 & 18.9 & \\
 & \tgtelmo{} & 13.3 & 16.4 & 14.1 & 16.2 & 18.8 & 14.9 & 21.3 & 16.4 & \\
 & \tgtft{} & 14.7 & 17.2 & 15.8 & 18.4 & 21.4 & 16.9 & 24.2 & 18.4 & \\
 & \srcelmosharedec{} & 17.4 & 20.8 & 18.6 & 21.5 & 24.9 & 20.3 & 29.0 & \bf 21.8 & \\
\midrule
\multirow{8}{*}{\shortstack{320K}}
 & baseline & 17.2 & 20.4 & 18.1 & 21.2 & 25.0 & 19.6 & 28.9 & 21.5 & \\
 & \shared{} & 18.1 & 21.1 & 19.1 & 22.4 & 26.3 & 21.2 & 30.6 & 22.7 & \\
 & \shared{}+bitext-BPE & 17.6 & 20.6 & 19.1 & 22.3 & 26.1 & 20.8 & 29.9 & 22.3 & \\
 & src-elmo & 18.8 & 22.3 & 21.1 & 24.0 & 27.5 & 22.2 & 32.5 & 24.1 & \\
 & \srcft{} & 19.0 & 22.5 & 20.9 & 23.5 & 26.9 & 22.2 & 32.1 & 23.9 & \\
 & \tgtelmo{} & 16.7 & 20.7 & 18.2 & 20.9 & 24.1 & 19.4 & 28.0 & 21.1 & \\
 & \tgtft{} & 16.1 & 19.4 & 17.1 & 20.0 & 23.1 & 18.5 & 26.3 & 20.1 & \\
 & \srcelmosharedec{} & 19.5 & 22.9 & 21.2 & 24.0 & 27.4 & 22.4 & 32.3 & \bf 24.2 & \\
\midrule
\multirow{8}{*}{\shortstack{640K}}
 & baseline & 19.2 & 22.9 & 21.2 & 24.5 & 27.9 & 22.4 & 33.1 & 24.5 & \\
 & \shared{} & 19.9 & 23.4 & 22.1 & 25.1 & 28.8 & 23.0 & 34.1 & 25.2 & \\
 & \shared{}+bitext-BPE & 19.4 & 22.8 & 21.7 & 24.9 & 28.4 & 22.9 & 33.6 & 24.8 & \\
 & src-elmo & 21.0 & 24.3 & 23.4 & 26.5 & 30.0 & 24.6 & 35.6 & 26.5 & \\
 & \srcft{} & 20.5 & 24.0 & 22.9 & 26.1 & 29.1 & 24.4 & 34.9 & 26.0 & \\
 & \tgtelmo{} & 18.9 & 22.6 & 20.8 & 24.2 & 27.5 & 22.3 & 31.9 & 24.0 & \\
 & \tgtft{} & 18.2 & 21.8 & 20.6 & 23.7 & 27.0 & 21.8 & 31.4 & 23.5 & \\
 & \srcelmosharedec{} & 21.2 & 25.1 & 23.9 & 26.7 & 30.2 & 24.7 & 36.2 & \bf 26.9 & \\
\midrule
\multirow{8}{*}{\shortstack{1280K}}
 & baseline & 20.9 & 24.6 & 23.6 & 26.5 & 30.5 & 24.7 & 36.2 & 26.7 & \\
 & \shared{} & 21.1 & 24.6 & 24.6 & 27.6 & 31.0 & 25.2 & 37.3 & 27.3 & \\
 & \shared{}+bitext-BPE & 20.5 & 24.0 & 23.9 & 26.2 & 30.6 & 24.8 & 36.2 & 26.6 & \\
 & src-elmo & 22.1 & 25.7 & 25.7 & 28.5 & 31.7 & 26.3 & 38.2 & 28.3 & \\
 & \srcft{} & 21.3 & 25.2 & 25.3 & 28.5 & 31.1 & 26.2 & 37.4 & 27.9 & \\
 & \tgtelmo{} & 20.9 & 24.4 & 23.6 & 26.6 & 30.3 & 24.9 & 36.1 & 26.7 & \\
 & \tgtft{} & 20.1 & 23.7 & 22.4 & 25.2 & 29.1 & 23.6 & 34.4 & 25.5 & \\
 & \srcelmosharedec{} & 22.3 & 26.0 & 26.3 & 28.9 & 32.6 & 26.8 & 38.6 & \bf 28.8 & \\
\midrule
\multirow{8}{*}{\shortstack{2560K}}
 & baseline & 21.7 & 25.6 & 25.4 & 28.2 & 32.3 & 26.2 & 39.1 & 28.4 & \\
 & \shared{} & 22.2 & 25.9 & 25.7 & 28.3 & 32.1 & 26.3 & 38.9 & 28.5 & \\
 & \shared{}+bitext-BPE & 21.8 & 25.5 & 25.5 & 27.9 & 32.1 & 26.0 & 38.6 & 28.2 & \\
 & src-elmo & 22.9 & 27.0 & 27.0 & 30.0 & 33.4 & 28.0 & 40.0 & \bf 29.8 & \\
 & \srcft{} & 22.2 & 26.4 & 26.3 & 29.5 & 32.4 & 27.3 & 39.3 & 29.1 & \\
 & \tgtelmo{} & 21.8 & 25.7 & 25.8 & 28.5 & 32.3 & 26.6 & 39.3 & 28.6 & \\
 & \tgtft{} & 21.5 & 25.3 & 24.5 & 27.0 & 30.2 & 25.2 & 36.8 & 27.2 & \\
 & \srcelmosharedec{} & 23.1 & 27.2 & 27.1 & 29.7 & 33.7 & 27.9 & 40.0 & \bf 29.8 & \\
\midrule
\multirow{8}{*}{\shortstack{5186K}}
 & baseline & 23.1 & 26.8 & 27.7 & 30.1 & 33.6 & 27.9 & 40.1 & 29.9 & \\
 & \shared{} & 22.6 & 26.6 & 27.7 & 30.5 & 33.4 & 28.2 & 40.2 & 29.9 & \\
 & \shared{}+bitext-BPE & 22.5 & 26.0 & 27.0 & 29.7 & 33.4 & 27.7 & 40.6 & 29.6 & \\
 & src-elmo & 23.7 & 27.8 & 28.7 & 31.1 & 34.5 & 29.2 & 41.8 & \bf 31.0 & \\
 & \srcft{} & 23.1 & 27.0 & 27.8 & 30.5 & 33.7 & 28.3 & 40.8 & 30.2 & \\
 & \tgtelmo{} & 22.9 & 26.6 & 26.9 & 29.5 & 33.8 & 27.7 & 40.5 & 29.7 & \\
 & \tgtft{} & 22.3 & 26.1 & 26.1 & 28.9 & 32.5 & 26.5 & 38.8 & 28.7 & \\
& \srcelmosharedec{} & 23.4 & 28.0 & 28.8 & 31.2 & 34.5 & 28.7 & 41.8 & 30.9 & \\
 \bottomrule
\end{tabular}
\caption{BLEU on newstest2012 to newstest2018 of WMT English-German translation in varioius simulated bitext size scenarios (cf. Figure~\ref{fig:data_news2017_delta_bar}).
}
\label{tab:mttestall}
\end{small}
\end{table*}

\clearpage
\section{Training and inference speed}
\label{app:speed}

\begin{table*}[h!]
\centering
\begin{tabular}{lrr}
\toprule
& train (tok/sec) & inference (tok/sec) \\
\midrule
\shared{} & 528,802 & 2,334 \\
\midrule
\srcelmo{} & 100,636 & 2,011 \\
\srcft{} & 57,753 & 2,080 \\
\tgtelmo{} & 142,525 & 259 \\
\tgtft{} & 95,313 & 299 \\
\bottomrule
\end{tabular}
\caption{Training and inference speed of models trained on WMT English-German. Training speed based on 32 V100 GPUs. Inference speed measured on a single V100 and by batching up to 12K source or target tokens.
}
\label{tab:speed}
\end{table*}

\end{document}